\documentclass{article}
\usepackage{ijcai25}

\usepackage{times}
\usepackage{soul}
\usepackage{url}
\usepackage[hidelinks]{hyperref}
\usepackage[utf8]{inputenc}
\usepackage[small]{caption}
\usepackage{graphicx}
\usepackage{amsmath}
\usepackage{amsthm}
\usepackage{booktabs}
\usepackage{algorithm}
\usepackage{algorithmic}
\usepackage[switch]{lineno}
\usepackage{bibentry}
\usepackage{amssymb}
\usepackage{color}

\title{Adaptive Prototype Model for Attribute-based Multi-label Few-shot Action Recognition}

\author{
    Juefeng Xiao, Tianqi Xiang, Zhigang Tu*
    \affiliations
    Wuhan University
    \emails
    \{juefengxiao, tianqixiang, tuzhigang\}@whu.edu.cn
}

\begin{document}

\maketitle

\begin{abstract}
In real-world action recognition systems, incorporating more attributes helps achieve a more comprehensive understanding of human behavior. However, using a single model to simultaneously recognize multiple attributes can lead to a decrease in accuracy. In this work, we propose a novel method \textit{i.e.} Adaptive Attribute Prototype Model (AAPM) for human action recognition, which captures rich action-relevant attribute information and strikes a balance between accuracy and robustness. Firstly, we introduce 
the Text-Constrain Module (TCM) to incorporate textual information from potential labels, and constrain the construction of different attributes prototype representations. 
In addition, we explore the Attribute Assignment Method (AAM) to address the issue of training bias and increase robustness during the training process.
Furthermore, we construct a new video dataset with attribute-based multi-label called Multi-Kinetics for evaluation, which contains various attribute labels (e.g. action, scene, object, etc.) related to human behavior. Extensive experiments demonstrate that our AAPM achieves the state-of-the-art performance in both attribute-based multi-label few-shot action recognition and single-label few-shot action recognition. The project and dataset are available at an anonymous account https://github.com/theAAPM/AAPM
\end{abstract}

\begin{figure}
\centering
\includegraphics[width=0.44\textwidth]{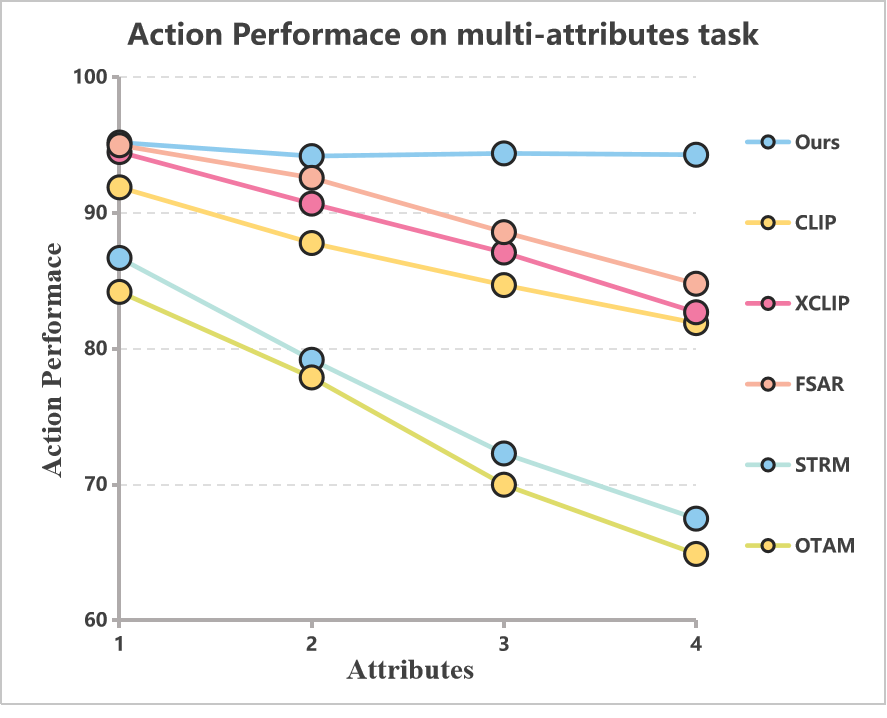}
\caption{The performance of different methods in constructing visual prototypes for action recognition when attributes increase. The results demonstrate that the performance of existing models will decrease as the number of attributes increases, while our AAPM overcomes the attribute-bias and maintains good performance.}
\label{dot}
\end{figure}

\section{Introduction}
Recognizing human action in videos is one of the most popular tasks in computer vision~\cite{carreira2017quo,lin2019tsm,herath2017going,kong2022human}. To achieve a better understanding of actions, more descriptive \textit{attributes*} beyond basic action categories are needed. However, all the existing methods are designed for the specific task of action recognition. As a result, these models may perform well in recognizing actions but tend to introduce bias in feature extraction, such as overemphasizing certain action-related features and neglecting other important attributes, which can reduce their generalization ability to more complex tasks. Figure ~\ref{dot} illustrates this situation. However, this bias may not be harmful in tasks that focus on a single attribute but can limit the model's performance across a wider range of attributes. We refer to this as attribute-bias.

\begin{quote}
    \textbf{Attribute*}: We define an attribute as a high-level category or abstract concept that encompasses a set of related subcategories, which share a common classification logic. For instance, 'action' is an attribute that includes various human activities such as playing soccer, swimming, dancing, etc. These subcategories can be further refined into more specific attributes at different levels of granularity. For example, 'dancing' can be considered an attribute that encompasses distinct dance styles like Latin dance, folk dance, ballet, etc.
    \label{attri}
\end{quote}

The VLP (Vision-Language Pretraining) models \cite{chen2023vlp} address the problem of attribute-bias through robust vision-language pre-training, which enhances the model's robustness and generalization ability. CLIP \cite{radford2021learning} aligns visual features with textual features by leveraging contrastive learning, enabling the model to be applied to the recognition of various attributes. The success of \cite{ni2022expanding,wang2021actionclip} on action recognition demonstrates that the VLP model has made remarkable progress on the downstream video tasks. However, to make the model perform comparably to other models in action recognition, these downstream tasks involve fine-tuning the VLP encoder. This process sacrifices the model's generalization ability, leading to bias towards specific tasks.
Additionally, semantic drift refers to the model overfitting to the semantic understanding of the training data. This can happen during the fine-tuning of the text encoder, especially when the text volume is limited.

On the other hand, few-shot learning \cite{fei2006one} addresses attribute-bias by constructing visual prototypes, enabling the model to generalize better with limited data. Based on this advantage, few-shot learning  \cite{vinyals2016matching,snell2017prototypical} has been widely applied in action recognition \cite{guo2020broader}. However, none of these few-shot action recognition methods has explored obtaining more action-related information, as they are constrained to the single attribute of action. An important reason is that when it comes to a multi-attribute way, the prototype representations may fall into confusion among various attributes, which will be discussed in detail in Section~\ref{prototypical confusion}. 

Meanwhile, multi-label classification tasks \cite{tsoumakas2007multi,chua2009nus,chen2019multi,xu2023learning} annotate multiple labels to the data to provide more detailed information. However, the previous methods lack flexibility, as they require continuously adding new classification heads or discriminators and retraining the model to accommodate a broader range of categories. They mainly concentrate on discriminating whether the objects appear in the image/video without taking into account extracting the advanced information accurately that we are interested in. Therefore, the prior multi-label methods are not suitable for human action recognition.

Based on the above observations, we exploit a method Adaptive Attribute Prototype Model (AAPM) for human action recognition, which can learn not only the human action but also other valuable action attribute information. Since AAPM leverages the textual information to provide guidance to construct different attributes of prototype representations, it enables the query data to recognize multiple attributes. To achieve this goal, we introduce Text-Constrain Module (TCM) to contribute robust prototype and exploit Attribute Assignment Method (AAM) for fair optimization. AAPM largely reduces the attribute-bias and enables the simultaneous recognition of multiple attributes.

TCM leverages both video and text information to construct prototypes. We utilize the CLIP model to extract the textual feature and visual feature of video and label text, where the textual feature represents the classification logic and the visual feature represents the pattern to be recognized. However, relying solely on text features to construct support sets is insufficient, as videos contain richer temporal features that can better describe the entire motion process. The module firstly learns the adaptive attribute feature via a transformer block by the text feature, and then constrains the visual feature through the cross attention block to ensure that each prototype is constructed under the definite attribute to avoid confusion.

However, the adaptive attribute feature in the latent space is hard to supervise. Besides, different attributes of a video may be relevant, leading to difficulty in extracting effective information by TCM. To overcome these problems, the AAM is adopted to assign stochastic irrelevant attributes to video data during the training process to make the optimization more robust. Additionally, by assigning more attributes, the text volume is much richer than previous works that only focus on  action class. As a result, our Attribute Assignment module largely reduces the issue of semantic drift and increases the model's generalizability.

To validate the effectiveness of our model, we construct the Multi-Kinetics dataset, which is a multi-attribute labels dataset. Different from the prior ones \cite{chua2009nus}, our dataset incorporates explicit classification logic among data, thus it provides a more structured and meaningful annotation framework. Meanwhile, this dataset introduces a new practical task called Attribute-based Multi-label Few-shot Action Recognition (AMFAR), which aims to recognize not only human action but also additional valuable information. Extensive experiments show that AAPM achieves the state-of-the-art performance on few-shot action recognition and has superior ability to represent multiple attributes, providing a benchmark for AMFAR.

In summary, our main contributions are listed below:
\begin{itemize}
    \item For the first time, we introduce the attribute-based mutlti-label action recognition to better reflect the practical demands of human action recognition in real-world. Meanwhile, we present a novel method AAPM for few-shot action recognition, which learns more helpful attributes information related to human action. 
    \item We design TCM to address the issues of attribute-bias and mitigate performance degradation on multiple attributes tasks. Furthermore, we propose a training paradigm AAM specifically for multi-attribute learning, which prevents the model from relying on the correlations between attributes.
    \item We construct a new video dataset with attribute-based multiple labels named Multi-Kinetics. Extensive experiments validate the effectiveness and generalizability of AAPM, where it obtains the outstanding performance in both the tasks of attribute-based multi-label few-shot action recognition and few-shot action recognition.
\end{itemize}

\begin{figure*}
\centering
\includegraphics[width=1\textwidth]{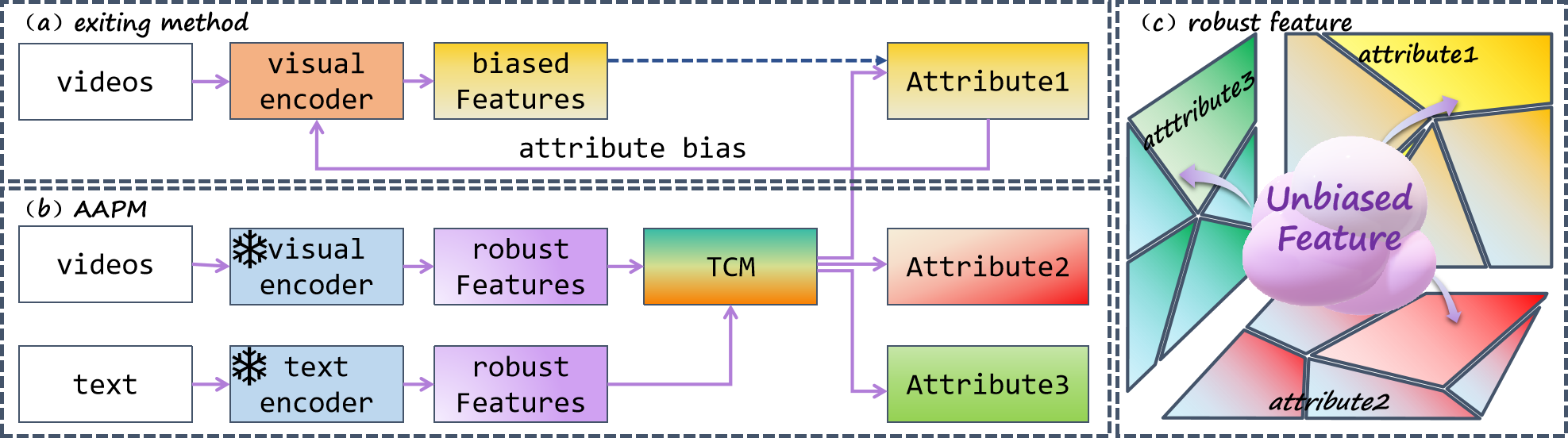}
\caption{(a) illustrates the recognition process of exiting methods. During the optimization, the training data introduces specific attribute bias into the feature encoder, causing the model fail to generalize to more attributes. (b) illustrates the recognition process of AAPM. AAPM freezes the visual and text encoders to ensure that the model can extract robust raw features, and then applies TCM to impose constraints on the target features. (c) illustrates the operating principle of TCM. The robust unbiased features containing rich information, and TCM maps the feature to multiple attributes.}
\label{motivation}
\end{figure*}

\section{Related Works}
\label{related}

\paragraph{Vision Language Pre-training.}

Recently, the Vision Language pre-training model has garnered significant attention due to its remarkable generalization ability and effectiveness. By training on billions of image-text pairs, \cite{radford2021learning,jia2021scaling} have achieved impressive results in the image-text matching task. Building upon its success, adaptations \cite{ni2022expanding,wang2021actionclip,xu2021videoclip} have been developed for Action Recognition, introducing methods for few-shot learning and zero-shot action recognition. However, these methods did not fully exploit the expressive power of the VLP model for text features. For better vision-text alignment, \cite{zhou2022learning,zhou2022conditional} try to learn the adaptive text prompt vectors to enhance the text feature. However, the starting point of these methods is flawed, since the text feature plays the role in defining the classification logic; a wiser idea is to use the text feature as a constraint for visual feature. More importantly, most existing downstream tasks \cite{wei2023improving,shen2021much} only fine-tune on the single attribute of action, compromising the generalization ability of the VLP model. During the fine-tuning process, feature encoder becomes more specialized and may result in semantic drift issues \cite{wu2023revisiting} for the text encoder. Our AAPM maximizes the generalizability of the VLP model through the proposed modules and training strategy, enabling the model to recognize a broader range of attributes.

\paragraph{Few-shot Action Recognition.}
Few-shot Learning \cite{fei2006one} aims to enable models to optimize from a small number of data, to classify the data in new classes. Due to the efficiency and effectiveness, it has been widely used in action recognition. The majority of existing few-shot recognition approaches \cite{snell2017prototypical,vinyals2016matching,sung2018learning,li2019finding} follow the metric-based manner to optimize the model and design robust alignment metrics to calculate the distances between the query and support data. 
Recently, a significant amount of researches \cite{cao2020few,perrett2021temporal,kong2022human,thatipelli2022spatio} have made progress on few-shot action recognition by leveraging temporal relations in video data. However, these methods have encountered a dilemma as they focus primarily on exploring the temporal dimension while neglecting the pursuit of richer information in the spatial dimension. 
One vital reason is that metric-based manner may cause prototypical confusion when dealing with multiple attributes. Since we are committed to provide more action-related information through few-shot action recognition, solving this confusion problem is inevitable.
On the other hand, \cite{rusu2018meta,rajeswaran2019meta,finn2017model} consider that a well-initialized model makes the fine-tuning process much easier; only a small number of gradient update steps are required to reach the optimum point. Since the CLIP model shows its strong ability in text-vision tasks, \cite{zhou2022learning,gao2024clip,wang2021actionclip,ni2022expanding} using the CLIP initialization to achieve good few-shot performance. But these methods are essentially not different from those of traditional action recognition; these fine-tuning-based methods still focus on a single attribute due to the generalization gap. In a word, there is no existing method that has leveraged the text information to enhance visual prototypes without compromising the generalization of the model. Therefore, our AAPM fills the gap in this field.

\paragraph{Multi-lablel Classification.}
Multi-label classification aims to provide more information by adding multiple labels. Standard multi-label classification tasks aim to predict a set of labels associated with a single data point. A conventional approach \cite{tsoumakas2007multi} involves training individual binary classifiers for each label present in the training dataset, without taking into account the dependencies among the labels.
Furthermore, \cite{chen2019multi,gong2013deep,zhu2017learning,sandouk2016multi} consider the correlation characteristics between labels and complete the learning of multi-label tasks through methods such as graph learning and structural learning. However, the labels of these methods are fixed and still require a significant amount of retraining when faced with newly introduced information. To overcome this problem, \cite{xu2023learning,wang2023clip} follow zero-shot methods, they use the VLP model to align visual and textual embedding to deal with unseen labels. But these methods only discriminate whether the label appears in the image/video and do not work with a standard recognition task. Since most existing multi-label datasets, such as \cite{chua2009nus}, lack explicit logic during the annotation process, only annotate different objects present in the images. As a result, the existing multi-label approaches based on these datasets have significant limitations \cite{yan2021deep,he2023open}, they can't precisely recognize the desired attributes to meet specific requirements. In this work, we create a attribute-based multi-label dataset Multi-Kinetics, introducing a more practical task AMFAR for multi-label methods.

\section{Methodology}

\subsection{Preliminary}
\paragraph{Attribute-based Multi-label Few-shot Action Recognition.}
As mentioned in the Introduction section, we introduce the new task AMFAR, and this section provides a detailed explanation to it. Assuming that we have a dataset $D$ with multi-attribute labels. A data contains multiple attributes $A_0, A_1, ..., A_m$, and each attribute has $l_0, l_1, ..., l_m $ different categories. For a given data $d$ in $D$, we have the following notations: $V_d $ represents the visual input, and $ L_{d}^i (i \in [0, m]) $ denotes the label under the classification attribute $A_i$.

The goal of AMFAR is to use a single model for classifying the attribute-based multi-label dataset. 
Above all, we follow the data partitioning scheme used in few-shot learning and create two non-overlapping datasets: a base training dataset $D_{train}$ and a testing dataset $D_{test}$. The labels in the two datasets are entirely different.

In the standard few-shot learning task, we typically sample numerous few-shot tasks from $D_{train}$ to optimize the model. In an N-way K-shot task, we construct a support set $D_{support} = \left\{ d_1, d_2, ..., d_N \times K \right\}$ consists of \textit{N} categories, with each category has \textit{K} videos. The objective of this task is to classify the query video \textit{q} in the support set.

Building upon this foundation, each video data has multiple attribute labels. Therefore, when constructing the support set, we constrain the \textit{N} categories sampled from the same attribute, resulting in $D_{support} = \left\{ d_1^i, d_2^i, ..., d_N^i \times K \right\}$, where $i \in [0, n]$ represents the attribute index. $C(D_{support})$ denotes the categories included in the support set, and $C(D_{support}^i)$ consists of $N(N<l_i)$ different categories belonging to the attribute $A_i$. The query set follows the same category sampling scheme, thus enabling the construction of prototypes under the equivalent classification logic.
 Compared to the previous few-shot action recognition, AMFAR faces more complex data formats, where each video data corresponds to multiple labels for different attributes. 
\begin{figure}
\centering
\includegraphics[width=0.45\textwidth]{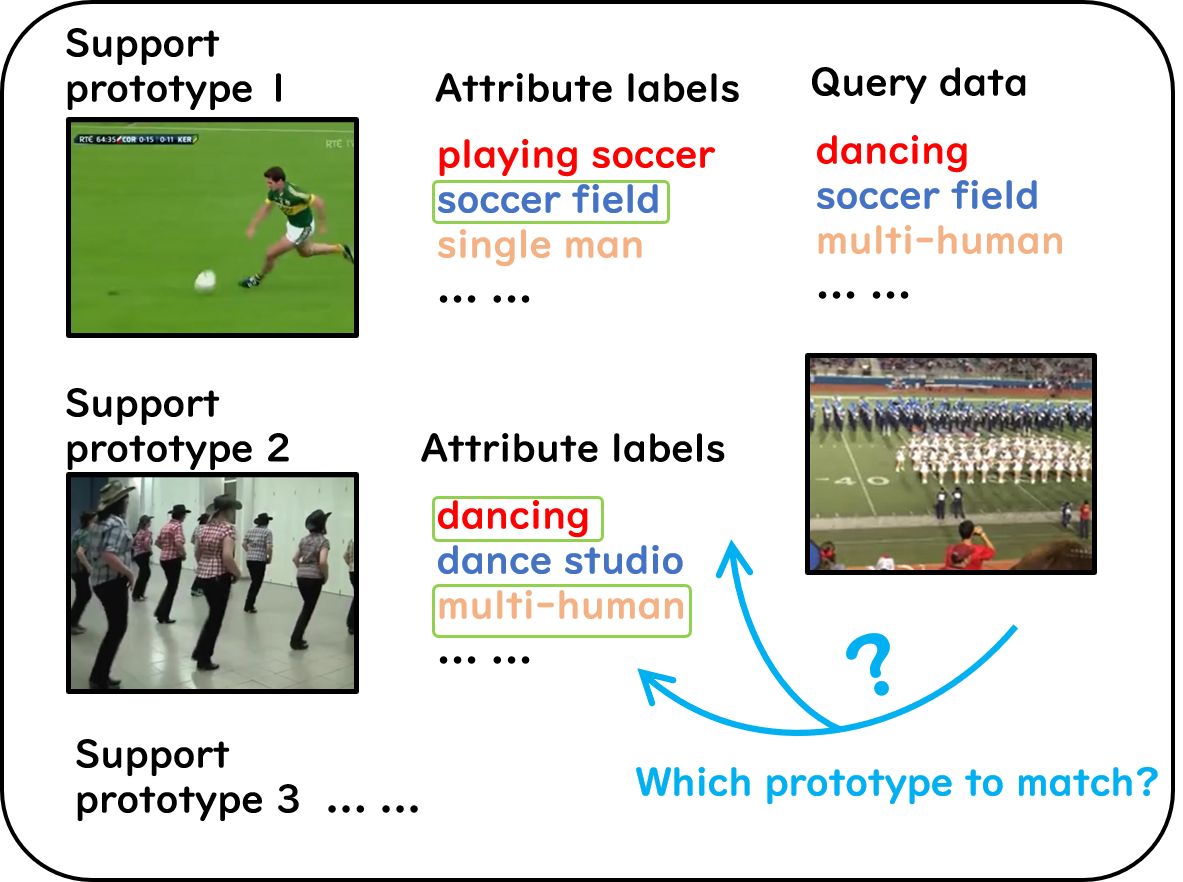}
\caption{The query data has the same scene attribute with prototype 1, and the same attributes of human group and action with prototype 2. The prototypical confusion occurs when the prior prototype networks applied to the multiple attributes recognition.}
\label{confusion}
\end{figure}
\paragraph{Prototypical Confusion.}
\label{prototypical confusion}
The phenomenon "confusion" is explained here, and the reason behind the poor performance of the prototype representations with multiple attributes is analyzed. Additionally, we present the concept and the definition of prototypical confusion.

Figure ~\ref{confusion} illustrates the dilemma faced by the prototype network. After constructing prototypes of support set, each support prototype contains multiple attributes such as action, scene, human group, etc. The problem arises when the query data attempts to match with the support prototypes, as it becomes challenging for the query data to determine which attribute is needed in the matching process. 
We refer to the confusion and ambiguity in the matching process, which is caused by query data and different support sets being the same on diverse attributes, as \textit{prototypical confusion}.

To handle this problem, the exiting works can be mainly classified into two ways. Firstly, these methods avoid the recognition of multiple attributes and only target on a specific attribute, where they use a data-driven way to automatically form a preference during the training process. However, it is evident that this way becomes ineffective when additional attributes are introduced, leading to a significant gap in this field. 
Secondly, they attempt to construct more powerful visual prototypes by increasing the number of shots. However, this way deviates from the essence of few-shot learning, as it requires a large amount of data to form robust features. Moreover, certain attributes exhibit strong correlations, such as action is often associated with scene. The association results in a loss of decoupling capability for multiple attributes recognition.
Obviously, these methods merely avoid the problem rather than addressing it fundamentally.

Notably, the prototypical confusion is not caused by AMFAR, but objectively exists in the prototype network. However, the previous tasks ignored this problem, causing it has not being explicitly expressed. Consequently, it is imperative to find an effective solution to address this issue.

\subsection{Text-Constrain module}

\begin{figure*}
\centering
\includegraphics[width=0.9\textwidth]{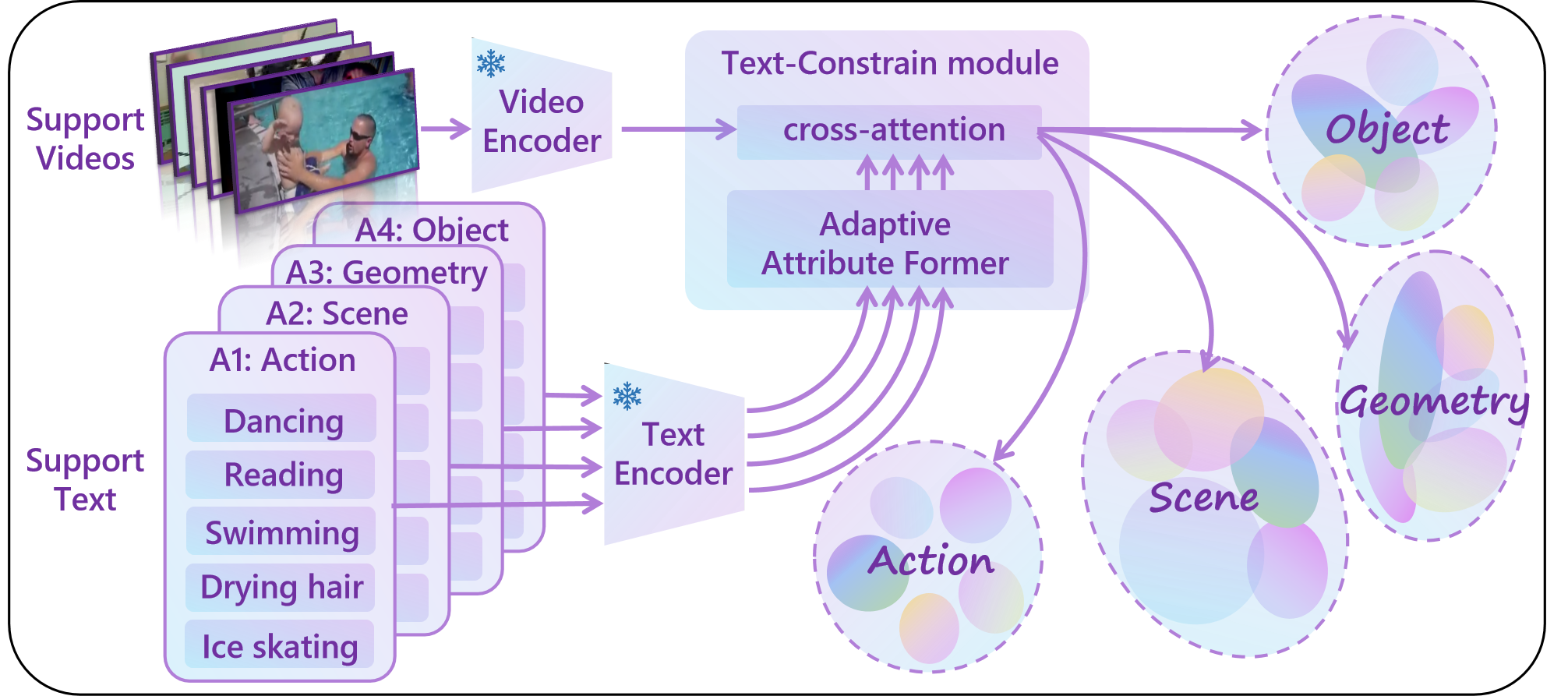}
\caption{The process of constructing visual prototypes via the support set. Firstly, the support videos are fed into the fixed video encoder to obtain video features, meanwhile the  text of support labels are fed into the text encoder to obtain text features. Secondly, the textual features of different attributes are individually fed into the adaptive attribute block to learn respective attribute spaces. Finally, the attribute spaces constrain the visual features by the cross-attention block, and construct the support prototypes of different attributes.}
\label{overview}
\end{figure*}

Figure ~\ref{overview} illustrates the process how the text information helps to constrain the attribute and build the support prototypes. For an input data $d_j\in D$, we use the frozen CLIP model to obtain a $t+1$ dimension feature, which contains textual feature $F_{Td_j}^i$ and visual features $ F_{Vd_j}=\{F_{Vd_j}^0, F_{Vd_j}^1, ..., F_{Vd_j}^{t-1}\}$, where $t$ represents the input video frames. After the attribute-based sampling, we construct the prototype representations for attribute $A_i$. However, if directly using the visual features to construct the prototypes, the classification logic of different attribute support sets cannot be unified, resulting in confusion during the model training. Therefore, we propose the Text-Constrain module that utilizes text features of labels to constrain the attribute.

\paragraph{Adaptive Attribute Space.} Firstly, the text features of the categories included in the support set $D_{support}^i$ are sent through the adaptive attribute block $\mathcal{G(\cdot)}$ to obtain a latent attribute space $\lambda_p$ specific to the support set $D_{support}^i$, and $\mathcal{G(\cdot)}$ is set as a trainable transformer encoder. 
\begin{equation}
    \label{eq1}
    \lambda_p(F_{Td}^i) = \mathcal{G}(C(D_{support}^i))
\end{equation}
From this point on, $\lambda_p$ contains the comprehensive text information of this support set, and then we use it to constrain the video features.

\paragraph{Cross Attention Constrain.} We then introduce a cross-attention mechanism, where the latent attribute space $\lambda_p$ provides the query vector $Q_p$, and the visual features $F_{Vd_j}$ provide key vector $K_v$ and value vector $V_v$. By leveraging the semantic information of the query, the visual features are constrained with the specific attribute:
\begin{equation}
    f_c(F_{Td^i},F_{Vd}) = softmax(Q_pK_v/\sqrt{d_k})V_v
\end{equation}
Where $d_k$ is the dimension of $K_v$, and $Q_p, K_v, V_V$ are obtained through a projection matrix $W$.
\paragraph{Prototype Alignment Distance.}  We obtain the constrained features of both support set and query set since they share the same latent attribute space $\lambda_p$. The next step is to calculate the prototypical distance. In this step, we directly follow the previous approach \cite{cao2020few,luo2021coco}, where a variant of dynamic time warping function $P$ is used to measure the distance $dist_{n}$ between the query data $d_c$ and the prototype representations $d_{s}^{in}, (n\in [0, N])$:
\begin{equation}
    dist_{n}(d_c, d_{s}^{in}) = P(f_{cs}^{in}, f_{cq}) 
\end{equation}
Where $f_{cs}^{in}$ represents the constrained support prototype of category n in Attribute i and $f_{cq}$ represents the constrained feature of query data. By calculating the distances between the average of support set features and the query feature, we can obtain the probabilities of the query data being assigned to different categories. Then, we use the cross-entropy to calculate the model loss.

\subsection{Attribute Assignment Method}

During the fine-tuning process of the VLP model, semantic drift can easily occur, especially with small datasets. To address this issue, we freeze both the image encoder and the text encoder of CLIP-ViT-B during training. However, homogeneous labels, which result in limited text variation, may still lead to this problem in subsequent modules. Therefore, we require a robust dataset with a sufficient number of attributes to train the model.

In practice, it is challenging to find a comprehensive dataset where each data point has rich attributes and follows a well-defined distribution. Most previous datasets focus on collecting and annotating data based on a single attribute. As a result, a dataset that covers multiple attributes is needed. Additionally, attributes in natural data are often correlated, which reduces the significance of text features during optimization and may even lead to semantic drift.

In light of this, we manually introduce some irrelevant attributes into the dataset during the training process, by inserting random shapes, color blocks and various objects to the video data. The assignment of each attribute has the independent probabilities. Taking the Kinetics-400 \cite{carreira2017quo} dataset as the baseline, we firstly annotate the scene, human group and illuminate information which are action-related in the video data. Then, we randomly assign various attributes to the data with the probabilities, and we use $P_0,..., P_{n}$ to represent the probabilities of all attributes.

\section{Experiments}
\subsection{Experiments Setup}
\paragraph{Datasets.}

As mentioned earlier, there is currently a lack of multi-label datasets based on different classification logic. Therefore, we have constructed a new dataset Multi-Kinetic. 
Following previous few-shot learning method \cite{zhu2018compound} on Kinetics-400 \cite{carreira2017quo}, we randomly selected 100 action categories and each category consists of 100 randomly sampled videos. Then we further do the annotation work and assignment method for the sampled dataset. Firstly, we annotate the attribute of scene, human group and illumination. They are divided into 34, 4 and 2 categories respectively. And then we introduce two additional attributes to the data using a random Attribute Assignment method, where Geometry attribute contains 65 categories of colorful geometry mask and Object attribute contains 15 categories of different objects mask. 
For the evaluation, the we choose 4 attributes with numerous categories \textit{i.e.} action, scene, geometry and object, since the attributes with the categories less than 5 \textit{i.e.} illumination, human group are not suitable for the standard 5-way few-shot experiment. For action attribute, we set 64 categories for training, 12 categories for validation and 24 categories for testing, similarly, we do the split of 19/5/10 on scene attribute, 41/8/16 on geometry attribute and 5/5/5 on object attribute. 
  

\paragraph{Inplementation Details.}

The backbone is implemented with
the CLIP-ViT-B/16 \cite{radford2021learning}, and 8 frames in resolution of 224×224 are sparsely sampled from each video, and we follow \cite{cao2020few} to measure the distance of prototypes. During the training process, the whole CLIP backbone is frozen. And we use AdamW optimizer with base learning rate
of 0.0001 and weight decay of 0.0005. For inference,
the average accuracy of 10,000 randomly sampled few-shot
tasks is reported in our experiments. And we
report the performance of our method under the 5-way
K-shot setting with K = 1 and K = 5 on each attribute. All models are trained on 4 NVIDIA 3090 GPUs and the batch size of support set are set to 1 and the batch size of query data is set to 25, the model has 89.3 million parameters and requires 5200 of memory per GPU.

\subsection{Attibute-based Multi-label Action Recognition}
    
\begin{table*}[h]
  \caption{Few-shot Multi-Attribute Classification}
  \label{table2}
    \begin{tabular}{p{0.08\textwidth}*{8}{p{0.08\textwidth}}}
      \hline
      \multicolumn{1}{c}{Method} & \multicolumn{2}{c}{Action} & \multicolumn{2}{c}{Scene} & \multicolumn{2}{c}{Geometry} & \multicolumn{2}{c}{Object}  \\
      \cline{2-9}
      & 1-shot & 5-shot & 1-shot & 5-shot & 1-shot & 5-shot & 1-shot & 5-shot \\
      \hline
      CLIP & 78.9 & 91.9 & 78.1 & 93.7 & 36.5  & 60.8 & 72.7 & 97.4\\
      AAPM & \textbf{90.8} & \textbf{94.8} & \textbf{86.6}  & \textbf{95.6} & \textbf{78.1} & \textbf{93.3} & \textbf{84.6} & \textbf{97.6} \\

      \hline
    \end{tabular}
\end{table*}
As mentioned earlier, our AAPM is driven by practical requirements and is groundbreaking in the task. As no similar work has been done before, we establish a new baseline by selecting the frozen visual encoder of CLIP-ViT-B to encode the feature, aiming to retain the sensitivity of the feature extraction layer to various attribute features. We evaluated the model on 5-way K-shot (K=1,5) few-shot tasks with the extracted visual features. Prior to the experiments, we set $P_0=1, P_1=P_2=P3=0.5$, ensuring that each data has 1 to 4 labels and these attributes are evaluated simultaneously.

The experimental results in Table ~\ref{table2} demonstrate that our method achieved significantly higher accuracy than the baseline on all four attributes. In the case of the Action attribute, we observed improvements of $11.9\%$ and $2.9\%$  in the 1-shot and 5-shot tasks, respectively. And while in Scene, Geometry, Object, the improvements are $8.5\%$, $41.6\%$, $11.9\%$ in 1-shot way and $1.9\%$, $32.5\%$, $0.2\%$ in 5-shot way. It shows that the improvement in 1-shot is more substantial for each attribute, while relatively smaller for the 5-shot tasks. This is because as K increases, the prototypes gradually learns the visual commonalities in the support set, thereby reducing classification error. The results indicate the superiority of our AAPM effectively utilizes text features to constrain visual prototypes.
It is worth mentioning that the number of parameters in our model is 89.3 million, which is only about $3\%$ higher than the 86.2 million in CLIP ViT-B/16.
Notably, there is a leap improvement on geometry attribute, 
this result is explainable and beneficial. Since geometry attribute is relatively low-level features information, such as "white rectangle", it can appear in videos that are not assigned with this attribute, thus it makes a great confusion during the recognition even the K increases.

\subsection{Few-shot Action Recognition}
\begin{table}[h]
  \caption{Comparision to the state-of-art method of Few-shot Action Recognition}
  \label{table3}
  \centering
    \begin{tabular}{c*{4}{c}c}
      \hline
      Method & Pre-training & Freeze & 1-shot & 5-shot\\
      \hline
      OTAM  & IN-1k & $\times$ & 72.2 & 84.2 \\
      STRM  & IN-1k & $\times$ & 62.9 & 86.7 \\
      HyRSM  & IN-1k & $\times$ & 73.7 & 86.1 \\
      AMeFu-Net  & IN-1k & $\times$ & 74.1 & 86.8 \\
      CLIP & CLIP-400M & $\times$ & 88.2 & 94.7 \\
      CLIP-FSAR  & CLIP-400M & $\times$ & 89.7 & 95.0 \\
      NameTuning  & CLIP-400M & $\times$ & 88.4 & 94.7 \\
      
      \hline
      CLIP & CLIP-400M & $\checkmark$ & 78.9 & 91.9 \\
      AAPM & CLIP-400M & $\checkmark $ & 90.8 & 94.8 \\
      AAPM* & CLIP-400M & $\checkmark$ & \textbf{91.5} & \textbf{95.2} \\

      \hline
    \end{tabular}
\end{table}
Since previous methods only compare single attribute of action, our AAPM further demonstrates its effectiveness by applying it to single few-shot Action Recognition task. We followed the setup in FSAR and conducted few-shot learning tasks on Kinetics-400 dataset. The experimental results indicate that our method achieves state-of-the-art (SOTA) performance in few-shot action recognition.

Based on the results in Table ~\ref{table3}, firstly, the "Freeze "means freezing the pre-trained encoder. By comparing the results of CLIP model with and without fine-tuning, there is an increase of $10.7\%$ and $2.9\% $in 1-shot and 5-shot tasks due to the fine-tuning. This demonstrates that, similar to most downstream work based on CLIP, the fine-tuning process improves accuracy on this task. But our AAPM retains the powerful general feature extraction capability of the CLIP model without fine-tuning the encoder.
Building upon this foundation, our method still achieves the improvements of $1.8\%$ and $0.2\%$ in 1-shot and 5-shot tasks, respectively. In "AAPM*" we adjust the Attribute Assignment method differently from Multi-Attribute Labels Classification by optimizing the value of P to achieve the best results. On the other hand, the "AAPM" still follows the settings of Multi-Attribute Labels Classification, where the model possesses both multi-label prediction capability and accuracy comparable to the SOTA.

\subsection{Ablation Study}
\label{ablation}

\paragraph{Attribute Assignment Method.}
\begin{table}[ht]
    \renewcommand{\arraystretch}{0.95}
    \centering
    \label{table4}
    \caption{Ablation study of Attribute Assignment}
    \begin{tabular}{c*{5}c}
    \hline
    \multicolumn{1}{c}{Assignment} & \multicolumn{2}{c}{Action} & \multicolumn{2}{c}{Object}  \\
    \cline{2-5}
       & 1-shot & 5-shot & 1-shot & 5-shot\\
      \hline
      $P_1=0, P_2=0$ & 91.4 & 94.8 & 70.8 & 95.0 \\
      $P_1=0.5, P_2=0$ & 91.5 & 94.7 & 69.8 & 94.4 \\
      $P_1=0, P_2=0.5$ & 91.4 & 95.1 & 82.1 & 96.4 \\
      $P_1=0.5, P_2=0.5$ & 91.3 & 94.9 & 81.8 & 96.4 \\
      \hline
    \end{tabular}
\end{table}

We adjust different Attribute Assignment strategies to explore the balance between model generalization and specialization. In this experiment, we ensure that the Action attribute is always included in the training while the Object attribute is always not $( P_0=1, P_3=0 )$. By adjusting the values of $ P_2 $ and  $P_3$, we control the degree of Attribute Assignment during training. The accuracy of the Action attribute represents specialization, while the accuracy of the Object attribute represents generalization.

We observe that the model always performs well on the seen attribute of action, exhibiting high accuracy. However, when $ P_1=P_2=0$, which means that no Attribute Assignment is applied, the model almost completely loses its generalization ability. It leads to serious semantic drift during training, resulting in poor performance on the unseen attribute Object, and the accuracy further decreases as training progresses.
As we set $P_1=0.5, P_2=0$, the performance on object attribute remains weak, but when we set $P_1=0,P_2=0.5$, the model performs well on unseen attribute categories and stabilizes as training progresses, demonstrating its generalization capability. The reason is that scene attribute represented by $P_1$ has a strong correlation with action attribute, and if only using these two attributes, the confusion still exists and may further decrease the model's generalization. On the other hand, the assigned attribute geometry represented by $P_2$ is unrelated to action attribute, it helps to enhance the model's generalization capability.
And when $P_1=P_2=0.5$, as more attributes are assigned, the model shows good performance in both specialization and generalization, the accuracy maintains at a high level while the attributes increase. In conclusion, Attribute Assignment module plays a positive role in improving the generalizability of the model.

\begin{table}[ht]
    \centering
    
    \caption{Ablation study of Text-Constrain module}
    \label{table5}
    \begin{tabular}{c*{5}c}
    \hline
    \multicolumn{1}{c}{Assignment} & \multicolumn{2}{c}{Text-Concatenation} & \multicolumn{2}{c}{Text-Constrain}  \\
    \cline{2-5}
       & 1-shot & 5-shot & 1-shot & 5-shot\\
       \hline
      Action & 80.0(-10.8) & 92.2(-2.6) &90.8& 94.8\\
      Scene & 79.6(-7.0) & 93.7(-1.9) &86.6&95.6\\
      Geometry& 38.5(-29.6) & 64.5(-28.8) &78.1&93.3\\
      Object& 73.0(-11.6) & 97.5(-0.1) &84.6&97.6\\
      \hline
    \end{tabular}
\end{table}

\paragraph{Text-Constrain Module.}
In this section, we simplified the Text-Constrain module, but to ensure fair comparison, we introduced text features for the comparative methods. We concatenated the text features with the video features along the temporal dimension, ensuring that the input features to the model contain both video and text information. The experimental results are shown in Table ~\ref{table5}. By comparing with Table ~\ref{table2}, we observe that simple concatenation of text features does not work well, and the Text-Constrain module has a positive impact on the model's performance. It leads to improvements of 10.8\%, 7.0\%, 29.6\%, and 11.6\% in the 1-shot tasks for Action, Scene, Geometry, and Object, respectively, and in the 5-shot tasks, there are improvements of 2.6\%, 1.9\%, 28.8\%, and 0.1\%. These results demonstrate the effectiveness of the Text-Constrain module. 
Table ~\ref{table6} shows the performance of existing methods in constructing visual prototypes for action recognition while attributes increase. Attributes are progressively introduced in the sequence of Action, Scene, Geometry, and Object during the training process. The results indicate that although VLP models can mitigate model degradation to a certain extent compared to pure visual models, their generalization ability is still significantly affected. However, TCM greatly reduces the degradation of the model caused by attribute-bias.

\begin{table}[ht]
    \centering
    \caption{Performance degradation of exiting methods}
    \label{table6}
    \begin{tabular}{c*{6}c}
    \hline
    \multicolumn{1}{c}{Assignment} & \multicolumn{1}{c}{Pre-training} & \multicolumn{4}{c}{Attributes}  \\
    \cline{3-6}
       & & 1 & 2 & 3 & 4\\
       \hline
      STRM & IN-1k & 86.7 & 79.2 & 72.3 & 67.5\\
      OTAM & IN-1k & 84.2 &77.9&70.0&64.9\\
      CLIP & CLIP-400M & 91.9 & 87.8 & 84.7 &81.9\\
      CLIP-FSAR & CLIP-400M & 95.0&92.6&88.6&84.8\\
      XCLIP & CLIP-400M & 94.5 & 90.7 & 87.1 &82.7\\
      AAPM & CLIP-400M & 95.2&94.2&94.4&94.3\\
      \hline
    \end{tabular}
\end{table}

\section{Conclusion}
We propose a novel model AAPM for few-shot action recognition, which achieves distinguishing not only the human action category but also more action-related attributes. Particularly, we propose the TCM to effectively address the problem of attribute-bias in multi-attribute tasks. TCM enables the model to extract robust and unbiased features, and incorporates the textual information of label text to constrain the construction of visual prototypes. To train the model, we design the AAM that balances the model's generalization and specialization, it avoids the negative impact of the correlation between attributes on the model's learning. In addition, we introduce a new dataset, termed Multi-Kinetics, for evaluation, which is the first multi-label human action dataset with multiple attributes. Remarkably, our model achieves the state-of-the-art performance in few-shot human action recognition with respect to both multi-label attributes prediction and single-label attribute prediction.

\appendix

\bibliographystyle{named}
\bibliography{ijcai25}

\end{document}